\documentclass{article}
\usepackage{float}
\usepackage[caption = false]{subfig}
\usepackage{spconf,amsmath,graphicx}
\usepackage{xcolor}
\usepackage{hyperref}
\usepackage{lipsum,multicol}
\usepackage{amsfonts}
\usepackage{ulem}

\graphicspath{ {img/} }

\title{Image storage on synthetic DNA using autoencoders}
\name{Xavier Pic$^{\star}$, Marc Antonini$^{\star}$}
\address{$^{\star}$ I3S laboratory, Côte d’Azur University and CNRS\\UMR 7271, Sophia Antipolis, France}
\begin{document}
%
\maketitle
\begin{abstract}
\vspace{-0.2\baselineskip}
\par

Over the past years, the ever-growing trend on data storage demand, more specifically for "cold" data (rarely accessed data), has motivated research for alternative systems of data storage.
Because of its biochemical characteristics, synthetic DNA molecules are now considered as serious candidates for this new kind of storage. This paper presents some results on lossy image compression methods based on convolutional autoencoders adapted to DNA data storage. 

The model architectures presented here have been designed to efficiently compress images, 
encode them into a quaternary code, and finally store them into synthetic DNA molecules. 
This work also aims at making the compression models better fit the problematics that we encounter when storing data into DNA, 
namely the fact that the DNA writing, storing and reading methods are error prone processes. 
The main take away of this kind of compressive autoencoder is our quantization and the robustness to substitution errors thanks to the noise model that we use during training.
\end{abstract}
\begin{keywords}
image, compression, DNA, autoencoder, denoising 
\end{keywords}
\vspace{-1\baselineskip}
\vspace{-0.5\baselineskip}
\section{Introduction}
\vspace{-0.5\baselineskip}
\label{sec:intro}
\par
Autoencoders has become a focus in the field of image compression. Quite early, some autoencoders \cite{Theis},\cite{Balle} and recurrent neural networks\cite{Toderici} were already showing better performances than the more conventional compression algorithms like JPEG or JPEG2000. Some research \cite{Fei}, \cite{Yoojin} aimed at having variable bit-rate compression autoencoders, other work put their focus on the differentiability problem of the quantization \cite{Theis}, \cite{dumas2018autoencoder} or the entropy evaluation \cite{Theis},\cite{oliveira}. Our goal is to adapt compression autoencoders to DNA data storage. For that, we have based our proposed network on the compression autoencoder designed by Lucas Theis et al. in \cite{Theis}, and introduced a DNA coder for the autoencoder's latent space tensor.
\vspace{-1\baselineskip}
\section{Context}
\vspace{-0.5\baselineskip}
\subsection{DNA data storage}
\vspace{-0.5\baselineskip}
\par 
The memory of humanity relies on our ability to manage increasingly large amounts of data, over periods of time ranging from a few years to several centuries. Current tools are no longer sufficient and it is necessary to consider game-changing solutions that can become operational quickly. One of the most promising solutions is to store information in the form of DNA, just like the genome by living beings. Indeed, DNA provides a very stable storage medium over very long periods with simple implementation conditions \cite{DNARAM}.
\par
The first step in the encoding workflow is the construction of a dictionary of codewords composed by the symbols A, G, C and T also called {\it nucleotides}. The DNA coded information stream must respect some biochemical constraints on the combinations of bases that form a DNA fragment: homopolymers, high/low GC content and repeated patterns should be avoided. One must also take into account that the process involves some biochemical procedures which can corrupt the encoded data. Synthesis, sequencing, storage and the manipulation of DNA (mainly PCR amplification) may introduce errors by introducing substitutions or indels (insertions or deletions of nucleotides), and may jeopardize the integrity of the stored content \cite{Goldman2013}.
\par
However, to ensure better adaptation to the characteristics of the storage medium, i.e., DNA, and possibly achieve higher storage efficiencies, it is better to design coding algorithms specific for DNA storage. Indeed, since DNA synthesis cost is relatively high, it is also important to take full advantage of an optimal compression that can be achieved before synthesizing the sequence into DNA. As an example, among other relevant works, \cite{DNAcoding} proposed Discrete Wavelet Transform (DWT) image decomposition where the DWT coefficients are scalar quantized and then encoded using a quaternary code. On the contrary, following the example of what is done today in the context of JPEG standardization for image coding with the development of JPEG AI\footnote{\color{blue}\url{https://jpeg.org/jpegai/index.html}}, we propose in this work to use neural networks such as autoencoders for image coding on quaternary codes.

\par
\vspace{-0.5\baselineskip}
\subsection{Compressive autoencoders for DNA image storage}
\vspace{-0.5\baselineskip}

\begin{figure*}[!ht]
    \centering
    \includegraphics[scale=0.5]{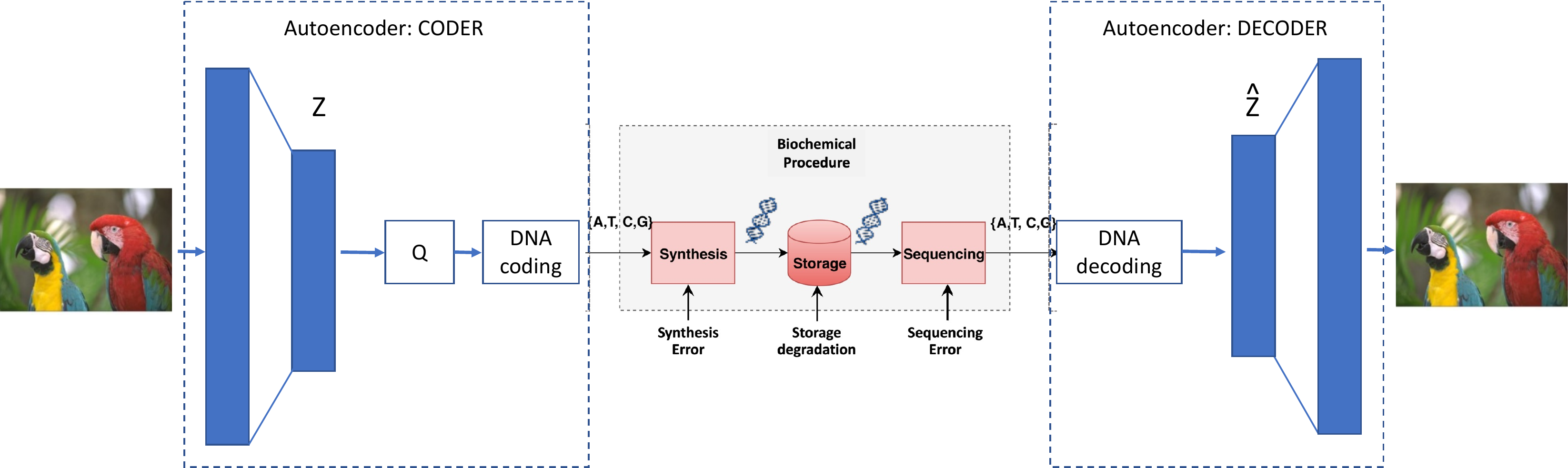}
    \vspace{-0.5\baselineskip}
    \caption{General scheme of an autoencoder used for image storage in synthetic DNA. $Z$ is the latent vector in the latent space. For compression purpose, $Z$ is quantized and encoded in a quaternary code for DNA storage. Then the quaternary code is used for DNA synthesis. The image reconstruction process needs sequencing before decoding with the autoencoder.}
    \label{fig:Auto}
\end{figure*}

The general idea of our proposed encoding process is depicted in the figure (\ref{fig:Auto}) and can be very roughly described by the following steps. Firstly, the input image has to be compressed using a lossy/near
lossless image coder. Here, we propose to use a compressive autoencoder to learn the image characteristics as well as the biochemical noisy process. Convolutionnal autoencoders have shown good properties for image denoising and could then appear as a good solution when it comes to deal with the noise introduced by the DNA data storage process. Secondly, a quantization operation is introduced to quantize the latent vector $Z$ in the latent space. The reason for that is that the latent space doesn't suffice as a compressed representation because it uses floating numbers. The quantization goal is to compress it and reduce its cost on the one hand and to facilitate its encoding on the other hand. Thirdly, in the case of DNA storage, the latent vector is encoded using a quaternary code using the $\{A,C,T,G\}$ alphabet. Decoding is ensured thanks to the decoder.
The quantization operation is one of the major problems when training compressive autoencoders. Indeed, to be able to train a model, all the operations have to be differentiable, which is not the case for the quantization (most of the times it is equivalent to a rounding function). However, this can be tackled by the use of a linear approximation function as in \cite{Theis}.
\par
During training, the compressive autoencoder has to minimize two quantities:
\vspace{-0.5\baselineskip}
\begin{itemize}
    \item The distortion which corresponds to the squared difference between the original image and the image reconstructed after compression and decompression.
    \vspace{-0.5\baselineskip}
    \item The entropy of the quantized latent space computed in a quaternary basis. The entropy being closely linked to the rate, we will use indifferently both terms in the rest of the paper.
\end{itemize} 
\vspace{-0.5\baselineskip}
The structure of the compressive autoencoder model we propose in this paper is closely linked the one proposed by Theis et al. in \cite{Theis}. It is described in the following section \ref{non-noisy}. The introduction of the biochemichal noise in the model is described in the section \ref{noisy}.

\par
\par
\vspace{-1\baselineskip}
\section{Non-noisy latent space}
\label{non-noisy}
\vspace{-0.5\baselineskip}
\subsection{Neural Network}
\vspace{-0.5\baselineskip}
\par
We designed our proposed autoencoder to obtain a latent space with higher dimensionality than the one proposed in \cite{Theis}, the objective of such a model was to obtain higher bit-rates.
Because we did not reduce the size of the latent space as much as in \cite{Theis}, our new model is shallower. This means that our new model now has a reduced number of parameters, which could cause some overfitting problems during the training process.
To compensate such a risk, we decided to add new residual blocks to the model with skipped connections.
\par
When encoding data, the common practice is to operate with integers, not with floating point numbers. For that reason, the output $Z$ of an autoencoder group of layers in the latent space cannot be encoded using quaternary code as it is and a quantization operator must then be introduced in the process. Given an input image $I$, the output image (decoded image) $\hat{I}$ of the autoencoder can be expressed as a combination of several functions such as follows:
\begin{equation}
    \hat{I} = g(\beta_{DNA}(\alpha_{DNA}(Q(f(I))))
\label{autofunc}
\end{equation}
where, $f$ represents the encoding part of the autoencoder and $g$ the decoding, $Q$ represents the quantizer that rounds the components z of the latent vector $Z=f(I)$ into integer values and is defined as:
\begin{equation}
\label{Quantizer}
    Q(z)=q\times\left \lfloor\frac{z}{q}+\frac{1}{2}\right \rfloor
\end{equation}
Finally, $\alpha_{DNA}$ and $\beta_{DNA}$ represent the DNA-encoding and decoding algorithms that encode a sequence of symbols into a quaternary stream composed by the letters of the alphabet $\{A,C,T,G\}$. When no noise, it is clear that $\beta_{DNA}(\alpha_{DNA}(k))=k \ \forall k \in \mathbb{Z}$. In this work we have used the DNA fixed-length code proposed in \cite{DNAcoding}.
\vspace{-0.5\baselineskip}
\subsection{Loss function}
\vspace{-0.5\baselineskip}
We used in our work the following classical loss function:
\begin{equation}
\label{Loss}
    L = ||I-\hat{I}||^2 + \lambda H(Q(Z)),
\end{equation}
where $I$ and $\hat{I}$ are respectively the input and output images and $H(Q(Z))$ corresponds to the entropy of the quantized latent space computed in base $4$ as follows:
\begin{equation}
    H(Q(Z))= -\sum_{i=1}^n Pr\{Q(z_i)\}log_4 Pr\{Q(z_i)\}
\end{equation}
with $Q(Z)=(Q(z_1),Q(z_2),...,Q(z_n))$ and $Pr\{Q(z_i)\}$ the probability of a quantized component. The entropy $H$ is expressed in {\it nucleotides per component}.
\par
To encode the quantized values, we use a fixed-length coding system.
Every value is coded with a DNA codeword of the same length $n$.
The set of all different codewords of length $n$ is a finite set.
In the case of quaternary code, there are $4^n$ different codewords, but only a subpart are DNA codewords that respect biochemical constraints (mainly no codes containing homopolymers runs such as AAAA or TTTT for example can be generated).
We call a codebook the set of constrained DNA codewords of length $n$ that can be used for this fixed-length encoding. Ensuring encodability of any compressed image means that the number of different possible values output by the quantizer has to be smaller or equal to the number of codewords available.
\par
The compression model output was bounded using a hyperbolic tangent function.
The uniform quantization was then applied to the output of that bounded function, giving a finite number of different possible values.
The number of possible values can be adjusted with the quantization step.
\vspace{-1\baselineskip}
\section{Introducing biochemical noise}
\label{noisy}
\vspace{-0.5\baselineskip}

As described previously, one must also take into account the process involves some biochemical procedures which can corrupt the encoded data. Synthesis, sequencing, storage and the manipulation of DNA may introduce errors by introducing substitutions or indels (insertions or deletions of nucleotides $A,T,C$ or $G$), and may jeopardize the integrity of the stored content \cite{Goldman2013}. DNA storage can then be viewed as a naturally noisy channel for which appropriately resilient encoding solutions need to be defined.
\par
\begin{figure}[h]
\centering
\includegraphics[scale=0.7]{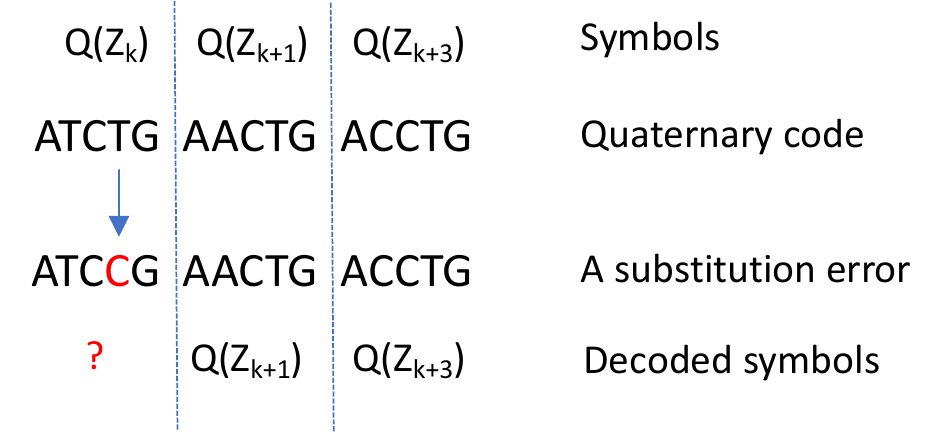}
\caption{Example of a substitution error on a DNA code. It only impacts one decoded symbol and not its neighbors.
}
\label{fig:Substitution}
\end{figure}

In order to adapt the autoencoders to the noisy channel of DNA data storage, one needs to introduce a noise model between the encoding and decoding parts. Since the substitution noise is prevalent in the noise of the DNA storage channel, we decided in this work to focus on it. A substitution, as mentioned earlier, is the phenomenon of one nucleotide being changed into another one (see figure \ref{fig:Substitution}).
\par
Furthermore, since we are using the fixed length codes of \cite{DNAcoding} for encoding the quantized values, a substitution error affects only one code and has no effect on the decoding of its neighbors as shown in figure \ref{fig:Substitution}. This means that the model for the substitution noise can be established and applied at the quantized tensor level (coefficients in the latent space) and not necessarily at the nucleotide level.\\
Let's call $\eta$ the substitution noise introduced by the biochemical process. One can rewrite the autoencoder input/output function given in equation (\ref{autofunc}) as:
\begin{equation}
    \hat{I} = g(\overline{\beta_{DNA}}(\alpha_{DNA}(Q(f(I)))+\eta)),
\label{autofuncNoise}
\end{equation}
where the noise is introduced at the level of the latent space and the quaternary decoder $\overline{\beta_{DNA}}$ is designed to decode a noisy code.
Because of the biochemical constraints, the DNA code is not a quaternary complete code and thus, a noisy code might not be decodable. To ensure decodability, when a noisy codeword is not decodable, we replace it with the closest valid possible code in the codebook, in terms of Hamming distance. 
\par
In this work we assumed that this noise can be modeled by a {\it i.i.d} Gaussian process.
The optimization of the autoencoder during training is still performed by minimizing the loss function given by equation (\ref{Loss}).

\par
\par
\par

\vspace{-1\baselineskip}
\section{Experimental results}
\vspace{-0.5\baselineskip}
\subsection{Implementation of the training}
\vspace{-0.5\baselineskip}
We trained the model with the 30k Flickr image dataset. 
During the training step, we used batches of 32 random crops of size 96x96 from the Flickr images. 
The training process has been separated into two steps: the first with a learning rate of 1e-4 during 200 epochs and the second one of 1e-5 that would be used for 500 epochs.
Models have been trained independently for each quantization step.
\par
The model was then evaluated using the Kodak dataset. For each compression rate chosen, we computed the performances for each image of the dataset, and also the average on the whole dataset. In figure \ref{fig:results_noised_retrained}, the gray curve (called avg) represents the average performance of the model and it's resistance to different levels of noise.

\begin{figure*}[ht]
\vspace{-0.5\baselineskip}
\centering
	\begin{minipage}[b]{0.4\linewidth}
		\centerline{
           \includegraphics[width = 1.2in]{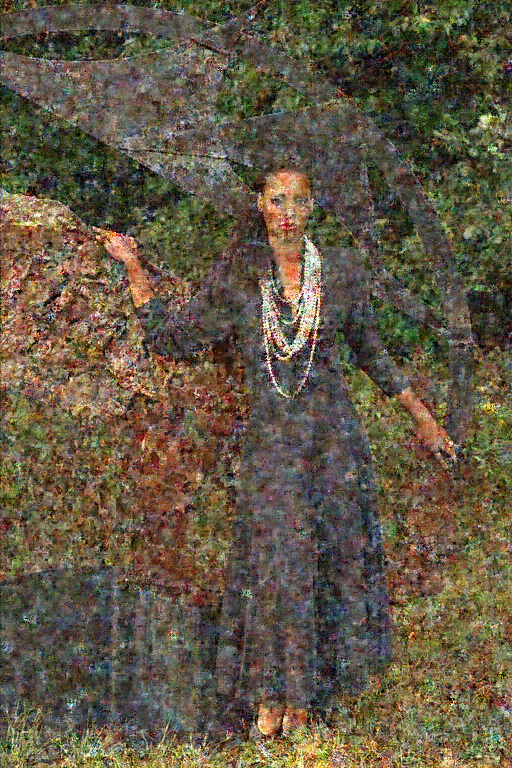}}
	   \end{minipage}
	\begin{minipage}[b]{0.4\linewidth}
		\centerline{
           \includegraphics[width = 1.2in]{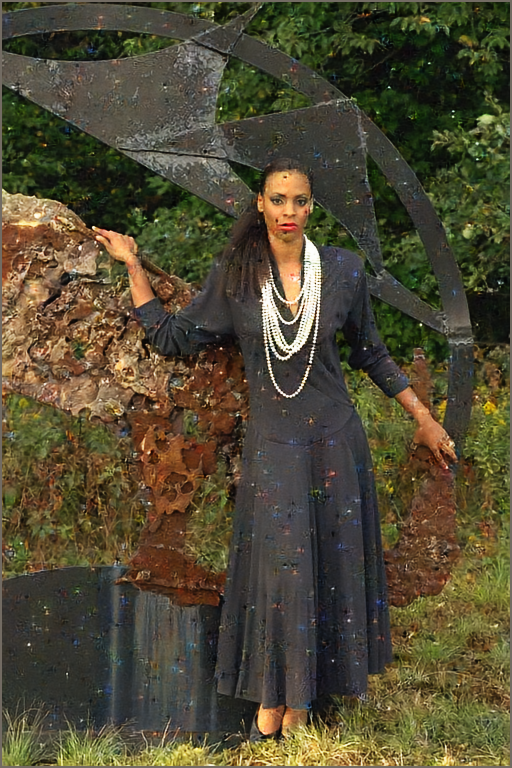}}
	   \end{minipage}
\centering
	\begin{minipage}[b]{0.4\linewidth}
		\centerline{
            \includegraphics[width = 1.8in]{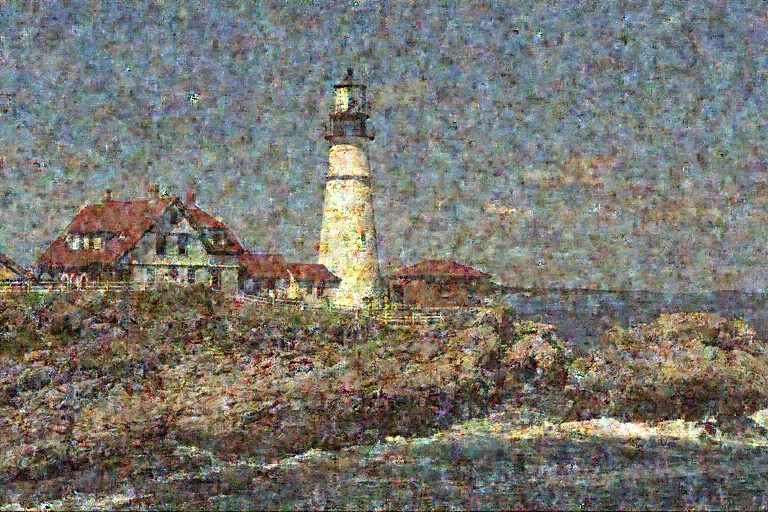}}
            \centering{(a) Without noise optimization}
	   \end{minipage}
	\begin{minipage}[b]{0.4\linewidth}
		\centerline{
            \includegraphics[width = 1.8in]{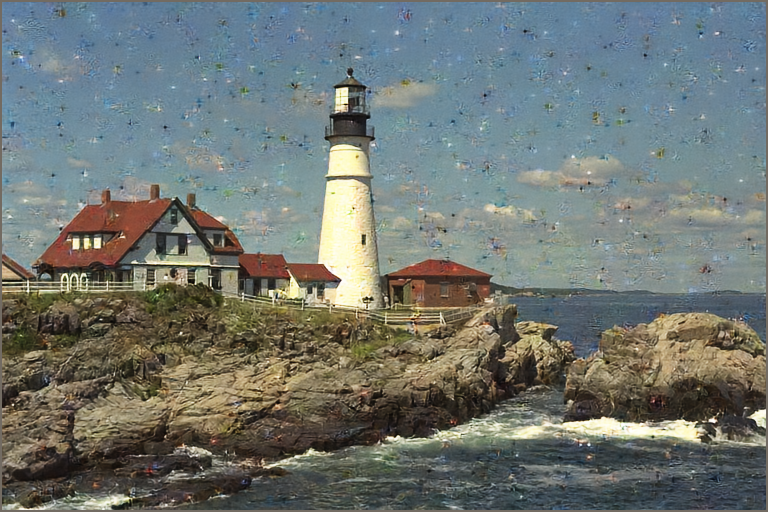}}
            \centering{(b) Optimized for noise}
	   \end{minipage}
\caption{
Visual results on kodim18 and kodim21 compressed at 6 bits/nt with noise levels of 5\%. Left: Model trained without any substitution noise considerations, right: training adapted to noise}
\label{fig:Visual}
\vspace{-0.5\baselineskip}
\end{figure*}
\vspace{-0.5\baselineskip}
\vspace{-0.5\baselineskip}
\subsection{Training process without substitution noise}
\vspace{-0.5\baselineskip}
 In this section, we describe the performance of the networks trained as described in section \ref{non-noisy} using the model of formula (\ref{autofunc}), and how their performance is maintained when noise is introduced into the channel at the encoding step and not during the training. The experiments were conducted on different autoencoder models, each one trained to a given compression rate (given quantization step $q$ as defined in formula (\ref{Quantizer})).

The result are presented in the figure \ref{fig:results_noised_retrained} where the PSNR is evaluated for different noise levels and averaged on all the images of the Kodak dataset. The results are provided for two different rates, 6 and 4 bits/nucleotide. 
The reconstruction maintains a good visual quality until the substitution noise level reaches around 5\%, where a lot of artifacts start to affect drastically the reconstruction, only maintaining an approximation of the general image features as shown in the figure \ref{fig:Visual}(a).
What we can quickly understand from those results is that the substitution noise has a big influence on the performances of our models. After a few percentages of error, the images technically become unusable. This justifies the development for compression methods adapted to noise.
\vspace{-0.7\baselineskip}
\subsection{Training process including substitution noise}
\vspace{-0.5\baselineskip}
In this section, we evaluate the performance of our autoencoder models trained with a noisy latent space as proposed in section \ref{noisy} and formula (\ref{autofuncNoise}). For each quantization step $q$ (or equivalently each rate in nucleotides per component) we trained a specific autoencoder.
\par
Here, the most important parameter is the noise level. It increases throughout the training from 0 (meaning that at the beginning of the training, the model is learning with a non-noisy data) and a value $max\_level$, which is the maximum noise level, used during the last epochs of the training session. 
Note that a more complex alternative would be to train several models each one for different noise levels, instead of training a unique model for all the possible noise levels.

\begin{figure}[ht]
    \centering
    \hspace{-0.25cm}
    \includegraphics[scale=0.35]{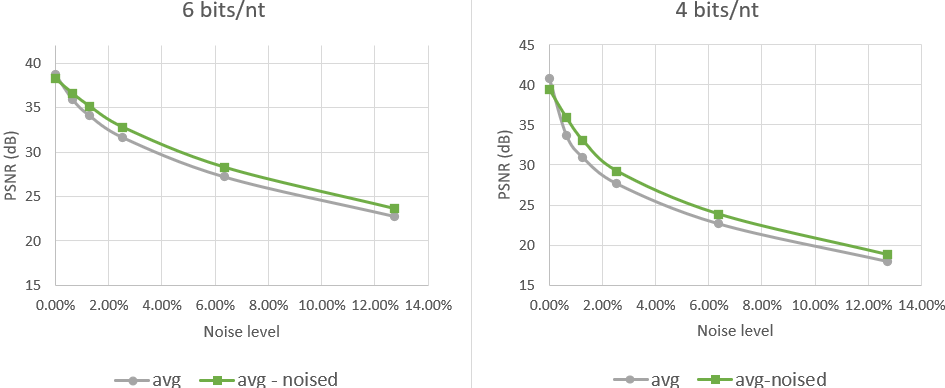}
    \caption{Comparison of robustness to noise at different coding rates for models trained without considering noise substitution (avg) and model trained by considering the substitution noise (avg-noise).}
    \label{fig:results_noised_retrained}
    \vspace{-0.5cm}
\end{figure}

What can be taken out from the results shown in the figure \ref{fig:results_noised_retrained} is that by training our models to adapt to a substitution noise we managed to obtain a gain between 0.5 and 1.5 dB on the PSNR depending on the level of noise. On the other hand, the models adapted to noise seem to underperform when no noise is introduced in the latent space. Furthermore, the visual results provided by the optimized autoencoder remain very interesting for a substitution noise level around 5\% (see figure \ref{fig:Visual}(b)) showing a strong robustness to high substitution noise levels and the good performance of the proposed solution.

\vspace{-1\baselineskip}
\section{Conclusion}
\vspace{-0.25\baselineskip}
In this work, we have developed a compression solution for image storage on synthetic DNA, robust to substitution noise. The proposed approach is based on compressive autoencoder optimized for DNA fixed-length encoding technologies.  
A noise model was developed and introduced in the autoencoder optimization to analyze its effects on the quality of reconstruction of the decoded image. Training the compression neural network including the noise model showed improvements over the network trained without the noise model (figure \ref{fig:results_noised_retrained}).
\par
In future works, experimenting with entropy-based DNA coding systems instead of fixed-length encoding might show some improvements since the compression network minimizes entropy. 
Introducing new types of noise (insertions and deletions), and solutions to minimize their effects is also another field of improvement.\color{black}

\end{document}